\documentclass[twocolumn,11pt]{article} 

\usepackage[utf8]{inputenc} 

\usepackage{geometry} 
\usepackage{graphicx} 

\usepackage{multirow}
\usepackage{booktabs} 
\usepackage{array} 
\usepackage{paralist} 
\usepackage{verbatim} 
\usepackage{lipsum}  
\usepackage{amsmath}
\usepackage{amsfonts}
\usepackage{hyperref}
\usepackage{abstract}
\usepackage{fancyhdr} 
\usepackage[nottoc,notlof,notlot]{tocbibind} 
\usepackage[titles,subfigure]{tocloft} 

\usepackage{subcaption}

\title{\bf A Multi-Task Learning Framework for Extracting Drugs and Their Interactions from Drug Labels}
\author{Tung Tran$^{1}$, Ramakanth Kavuluru$^{1,2}$, and Halil Kilicoglu$^{3}$}

\date{\footnotesize
$^{1}$Department of Computer Science, University of Kentucky, Lexington, KY, USA\\
$^{2}$Division of Biomedical Informatics, Department of Internal Medicine, University of Kentucky, Lexington, KY, USA\\
$^{3}$Lister Hill National Center for Biomedical Communications, National Library of Medicine, Bethesda, MD, USA
} 

\begin{document}
\maketitle
\global\csname @topnum\endcsname 0
\global\csname @botnum\endcsname 0
\begin{abstract}
Preventable adverse drug reactions as a result of medical errors present a growing concern in modern medicine. As drug-drug interactions (DDIs) may cause adverse reactions, being able to extracting DDIs from drug labels into machine-readable form is an important effort in effectively deploying drug safety information. The DDI track of TAC 2018 introduces two large hand-annotated test sets for the task of extracting DDIs from structured product labels with linkage to standard terminologies. Herein, we describe our approach to tackling tasks one and two of the DDI track, which corresponds to named entity recognition (NER) and sentence-level relation extraction respectively. Namely, our approach resembles a multi-task learning framework designed to jointly model various sub-tasks including NER and interaction type and outcome prediction. On NER, our system ranked second (among eight teams) at 33.00\% and 38.25\% F1 on Test Sets 1 and 2 respectively. On relation extraction, our system ranked second (among four teams) at 21.59\% and 23.55\% on Test Sets 1 and 2 respectively.
\end{abstract}

\section{Introduction}

Preventable adverse drug reactions (ADRs) introduce a growing concern in the modern healthcare system as they represent a large fraction of hospital admissions and play a significant role in increased health care costs~\cite{mcdonnell2002hospital}. Based on a study examining hospital admission data, it is estimated that approximately three to four percent of hospital admissions are caused by adverse events~\cite{brennan1991incidence}; moreover, it is estimated that between 53\% and 58\% of these events were due to medical errors~\cite{thomas1999costs} (and are  therefore considered preventable). Such preventable adverse events have been cited as the eighth leading cause of death in the U.S., with an estimated fatality rate of between 44,000 and 98,000 each year~\cite{donaldson2000err}. As drug-drug interactions (DDIs) may lead to preventable ADRs, being able to extract DDIs from structured product labeling (SPL) documents for prescription drugs is an important effort toward effective dissemination of drug safety information. The Text Analysis Conference (TAC) is a series of workshops aimed at  encouraging research in natural language processing (NLP) and related applications by providing large test collections along with a standard evaluation procedure. The Drug-Drug Interaction Extraction from Drug Labels track of TAC 2018~\cite{tac2018}, organized by the U.S. Food and Drug Administration (FDA) and U.S. National Library of Medicine (NLM), is established with the goal of transforming the contents of SPLs into a machine-readable format with linkage to standard terminologies.

\begin{figure}[t]
  \centering
  \includegraphics[width=\columnwidth]{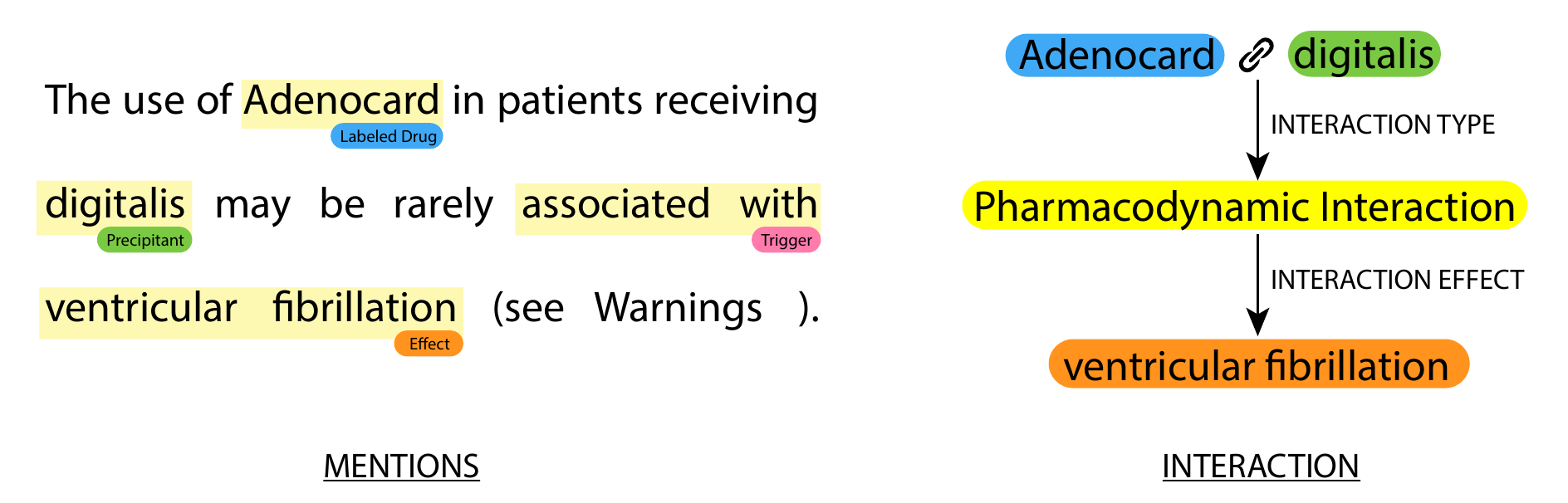}
  \caption{An example illustrating the DDI task}
  \label{fig_tac_example}
  \end{figure}

We focus on the first two tasks of the DDI track involving named entity recognition (NER) and relation extraction (RE). Task 1 is focused on identifying \textbf{mentions} in the text corresponding to precipitants, interaction triggers, and interaction effects. Precipitants are defined as substances, drugs, or a drug class involved in an interaction. Task 2 is focused on identifying sentence-level interactions; concretely, the goal is to identify the interacting precipitant, the type of the interaction, and outcome of the interaction. The interaction outcome depends on the interaction type as follows. Pharmacodynamic (PD) interactions are associated with a specified \emph{effect} corresponding to a span within the text that describes the outcome of the interaction. Naturally, it is possible for a precipitant to be involved in multiple PD interactions. Pharmacokinetic (PK) interactions are associated with a label from a fixed vocabulary of National Cancer Institute (NCI) Thesaurus codes indicating various levels of increase/decrease in functional measurements. For example, consider the sentence: ``There is evidence that treatment with phenytoin leads to to decrease intestinal absorption of furosemide, and consequently to lower peak serum furosemide concentrations.'' Here, \emph{phenytoin} is involved in a PK interaction with the label drug, \emph{furosemide}, and the type of PK interaction is indicated by the NCI Thesaurus code C54615 which describes a decrease in the maximum serum concentration (C$_{\text{max}}$) of the label drug. Lastly, \emph{unspecified} (UN) interactions  are interactions with an outcome that is not explicitly stated in the text and usually indicated through cautionary statements. Figure~\ref{fig_tac_example} features a simple example of a PD interaction that is extracted from the drug label for \textbf{Adenocard}, where the precipitant is \emph{digitalis} and the effect is ``ventricular fibrillation.''

\section{Materials and Methods}

Herein, we describe the training and testing data involved in this task and the metrics used for evaluation. In Section~\ref{sec-method}, we describe our modeling approach, our deep learning architecture, and our training procedure.

\subsection{Datasets}\label{sec-dataset}

\begin{table*}[t]\renewcommand{\arraystretch}{1.05}
  \centering
  \resizebox{\textwidth}{!}{
  \begin{tabular}{@{\extracolsep{1em}}l rrrr}
  \toprule
  & NLM-180 \textbf{*} & Training-22 & Test Set 1 & Test Set 2\\
  \midrule
  Number of Drug Labels & 180 & 22 & 57 & 66\\
  Mean number of sentences per Drug Label & 32 & 27 & 144 & 64\\
  Mean number of words per sentence & 23 & 24 & 22 & 23\\
  Proportion of \emph{annotated} sentences & 27\% & 51\% & 23\% & 23\%\\
  Mean number of mentions per \emph{annotated} sentence & 4.0 & 3.8 & 3.7 & 3.6\\
  \midrule
  Proportion of mentions that are Precipitant & 57\% & 53\% & 56\% & 55\%\\
  Proportion of mentions that are Trigger & 20\% & 28\% & 30\% & 33\%\\
  Proportion of mentions that are SpecificInteraction & 23\% & 19\% & 14\% & 12\%\\
  \midrule
  Proportion of interactions that are Pharmacodynamic  & 47\% & 49\% & 33\% & 28\%\\
  Proportion of interactions that are Pharmacokinetic & 25\% & 21\% & 28\% & 47\%\\
  Proportion of interactions that are Unspecified  & 28\% & 30\% & 39\% & 25\%\\
  \bottomrule
  \end{tabular}}
  \caption{Characteristics of datasets}
  \footnotesize \textbf{*} Statistics for NLM-180 were computed on mapped examples (based on our own\\ annotation mapping scheme) and not based on the original dataset.
  \label{tb_datastats}
  \end{table*}

Each drug label is a collection of sections (e.g., DOSAGE \& ADMINISTRATION, CONTRAINDICATIONS, and WARNINGS) where each section contains one or more sentences. Each sentence is annotated with a list of zero or more \emph{mentions} and \emph{interactions}. The training data released for this task contains 22 drug labels, referred to as Training-22, with gold standard annotations. Two test sets of 57 and 66 drug labels, referred to as Test Set 1 and 2 respectively, with gold standard annotations are used to evaluate participating systems. As Training-22 is a relatively small dataset, we additionally utilize an external dataset with 180 annotated drug labels dubbed NLM-180~\cite{nlm180} (more later). We provide summary statistics about these datasets in Table~\ref{tb_datastats}. Test Set 1 closely resembles Training-22 with respect to the sections that are annotated. However, Test Set 1 is more sparse in the sense that there are more sentences per drug label (144 vs. 27), with a smaller proportion of those sentences having gold annotations (23\% vs. 51\%). Test Set 2 is unique in that it contains annotations from only two sections, namely DRUG INTERACTIONS and CLINICAL PHARMACOLOGY, the latter of which is not represented in Training-22 (nor Test Set 1). Lastly, Training-22, Test Set 1, and Test Set 2 all vary with respect to the distribution of interaction types, with Training-22, Test Set 1, and Test Set 2 containing a higher proportion of PD, UN, and PK interactions respectively. 

\subsection{Evaluation Metrics}

We used the official evaluation metrics for NER and relation extraction based on the standard precision, recall, and F1 micro-averaged over exactly matched  entity/relation annotations. For either task, there are two matching criteria: \emph{primary} and \emph{relaxed}. For entity recognition, \emph{relaxed} matching considers only entity bounds while \emph{primary} matching considers entity bounds as well as the type of the entity. For relation extraction, \emph{relaxed} matching only considers precipitant drug (and their bounds) while \emph{primary} matching  comprehensively considers precipitant drugs and, for each, the corresponding interaction type and interaction outcome. As relation extraction evaluation takes into account the bounds of constituent entity predictions, relation extraction performance is heavily reliant on entity recognition performance. On the other hand, we note that while NER evaluation considers \emph{trigger} mentions, \emph{triggers} are ignored when evaluating relation extraction performance.

\subsection{Methodology}\label{sec-method}
We propose a multi-task learning framework for extracting drug-drug interactions from drug labels. The framework involves branching paths for each training objective (corresponding to sub-tasks) such that parameters of earlier layers (i.e., the context encoder) are shared.

\paragraph{Modeling Approach.} 
Since only drugs involved in an interaction (precipitants) are annotated in the ground truth, we model the task of precipitant recognition and interaction type prediction jointly. We accomplish this by reducing the problem to a sequence tagging problem via a novel NER tagging scheme. That is, for each precipitant drug, we additionally encode the associated interaction type. Hence, there are five possible tags: \textbf{T} for trigger, \textbf{E} for effects, and \textbf{D}, \textbf{K}, and \textbf{U} for precipitants with pharmaco\textbf{d}ynamic, pharmaco\textbf{k}inetic, and \textbf{u}nspecified interactions respectively. As a preprocesssing step, we identify the label drug in the sentence, if it is mentioned, and bind it to a generic entity token (e.g. ``LABELDRUG''). We additionally account for label drug aliases, such as the generic version of a brand-name drug, and bind them to the same entity token. Table~\ref{tb_example} shows how the tagging scheme is applied to the simple example in Figure~\ref{fig_tac_example}. A drawback is that simplifying assumptions must be made that will hamper recall; e.g., we only consider non-overlapping mentions (more later).

		\begin{table}[h]
		\renewcommand{\arraystretch}{1.2}
      \centering
			\resizebox{\columnwidth}{!}{
      \begin{tabular}{@{\extracolsep{1em}} cccccccccc }
        \toprule
        O & O & O & O & O & O & O & \textbf{B-D} \\
        The & use & of & LABELDRUG & in & patients & receiving & digitalis\\
        \midrule
        O & O & O & \textbf{B-T} & \textbf{I-T} & \textbf{B-E} & \textbf{I-E} & O\\
        may & be & rarely & associated & with & ventricular & fibrillation & .\\
        \bottomrule
      \end{tabular}}
      \caption{Example of the tagging scheme}
      \label{tb_example}
    \end{table}
    
Once we have identified the precipitant offsets (as well as of triggers/effects) and the interaction type for each precipitant, we subsequently predict the outcome or consequence of the interaction (if any). To that end, we consider all entity spans annotated with \textbf{K} tags and assign them a label from a static vocabulary of 20 NCI concept codes corresponding to PK consequence (i.e., multiclass classification) based on sentence-context. Likewise, we consider all entity spans annotated with \textbf{D} tags and link them to mention spans annotated with \textbf{E} tags; we accomplish this via binary classification of all pairwise combinations. For entity spans annotated with \textbf{U} tags, no outcome prediction is made.

\begin{figure*}[h]
		\centering
  		\includegraphics[width=\textwidth]{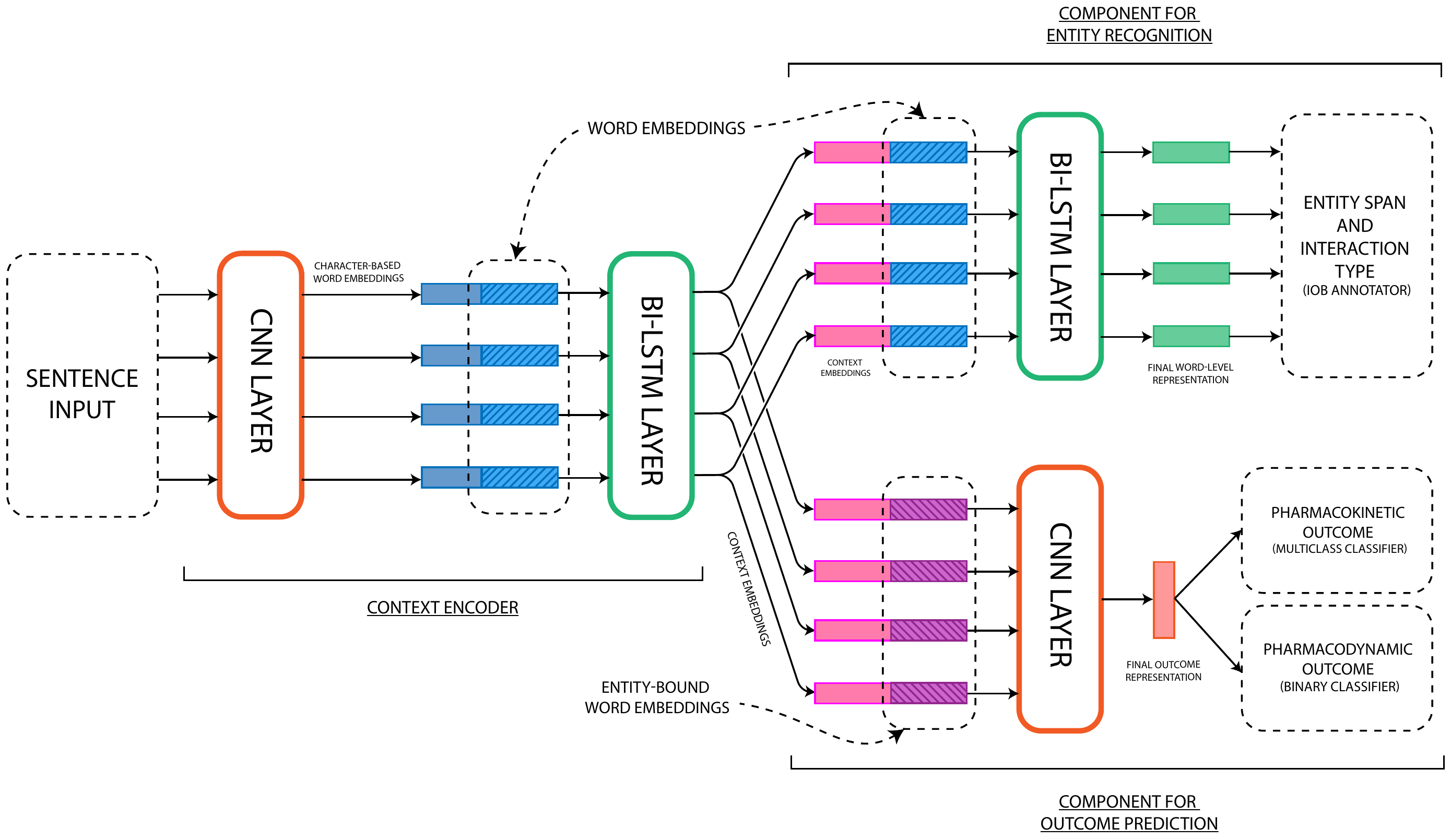}
		\caption{The multi-task neural network for DDI extraction}
		\label{fig_model}
    \end{figure*}
    
\paragraph{Neural Network Architecture.} 
Our proposed deep neural network is illustrated in Figure~\ref{fig_model}. We utilize Bi-directional Long Short-Term Memory networks (Bi-LSTMs) and convolutional neural networks (CNNs) designed for natural language processing as building blocks for our architecture~\cite{kimcnn,tran2017predicting}. Entity recognition and outcome prediction share common parameters via a Bi-LSTM context encoder that composes a context representation at each timestep based on input words mapped to dense embeddings and character-CNN composed representations. We use the same character-CNN representation as described in a prior work~\cite{tran2018end}; however, in this work, we omit the character type embedding. A Bi-LSTM component is used to annotate IOB tags for joint entity recognition and interaction type prediction (or, NER prediction) while a CNN with two separate dense output layers (one for PK and one for PD interactions) is used for outcome prediction. We consider NER prediction to be the main objective with outcome prediction playing a secondary role. When predicting outcome, the contextual input is arranged such that candidate entity (and effect) mentions are bound to generic tokens; the resulting representation is referred to as ``entity-bound word embeddings'' in Figure~\ref{fig_model}. 

\paragraph{Notation.} We denote $\text{BiLSTM}(\cdot) : \mathbb{R}^{n \times d_\text{in}} \mapsto \mathbb{R}^{n \times d_\text{out}}$ as an abstract function, representing a standard bi-directional recurrent neural network with LSTM units, where $n$ is the number of input vector representations (e.g., word embeddings) in the sequence and $d_\text{in}$ and $d_\text{out}$ are the dimensionality of the input and output representations respectively. We similarity denote $\text{CNN}^{[h_1, \ldots, h_k]}(\cdot) : \mathbb{R}^{n \times d_\text{in}} \mapsto \mathbb{R}^{d_\text{out}}$ to represent a standard CNN that maps an $n \times d_\text{in}$ matrix to a vector representation of length $d_\text{out}$, where $[h_1, \ldots, h_k]$ is a list of window (or kernel) sizes that are used in the convolution.

\paragraph{Context Encoder.}

Let the input be a sentence of length $n$ represented as a matrix $S \in \mathbb{R}^{n \times d}$, where each row corresponds to a word embedding of length $d$. Moreover, let $W^i \in \mathbb{R}^{m \times d_{char}}$ represent the word at position $i$ of the sentence such that each of the $m$ rows correspond to a character embedding of length $d_{char}$. The purpose of the context encoder is to encode each word of the input with surrounding linguistic features and long-distance dependency information. To that end, we employ the use of a Bi-LSTM network to encode S as a context matrix $C \in \mathbb{R}^{n \times d_{\text{context}}}$ where $d_{\text{context}}$ is a hyper-parameter of the network. Concretely, 
\begin{equation}
C = \text{BiLSTM}
\left(  {\begin{array}{*{20}c}
   S_1 \,\, \| \,\, \text{CNN}^{[3]}(W^1) \\
   \vdots\\
   S_n \,\, \| \,\, \text{CNN}^{[3]}(W^n) \\
 \end{array} }  \right)
\end{equation}
where $S_i$ denotes the $i^{\text{th}}$ row of $S$ and $\|$ is the vector concatenation operator. Essentially, for each word, we compose character representations using a CNN with a window size of three and concatenate them to pre-trained word embeddings; we stack the concatenated vectors as rows of a new matrix that is ultimately fed as input to the Bi-LSTM context encoder. The $i^{\text{th}}$ row of $C$, denoted as $C_i$, represents the entire context centered at the $i^{\text{th}}$ word. As an implementation detail, we chose $n$ and $m$ to be the maximum sentence and word length (according to the training data) respectively and pad shorter examples with \emph{zero} vectors.

\paragraph{NER Objective.}

The network for the NER objective manifests as a stacked Bi-LSTM architecture when we consider both the context encoder and the entity recognition component. Borrowing from residual networks~\cite{he2016deep}, we re-inforce the input by concatenating word embeddings to the intermediate context vectors before feeding it to the second Bi-LSTM layer. Concretely, the final entity recognition matrix $R \in \mathbb{R}^{n \times d_{\text{ner}}}$ is composed such that 
\begin{equation}
R = \text{BiLSTM}
\left(  {\begin{array}{*{20}c}
   C_1 \,\, \| \,\, S_1 \\
   \vdots\\
   C_n \,\, \| \,\, S_n \\
 \end{array} }  \right).
\end{equation}
The output at each position $i=1,\ldots, n$ is 
\begin{equation*}
\mathbf{q}^i = W^{\text{ner}} R_i + \mathbf{b}^{\text{ner}}
\end{equation*}
where $R_i$ is the $i^{\text{th}}$ row of $R$ and $W^{\text{ner}} \in \mathbb{R}^{\ell \times d_{\text{ner}}}$ and $\mathbf{b}^{\text{ner}} \in \mathbb{R}^{\ell}$ are network parameters such that $\ell = 11$ denotes the number of possible IOB tags such as \textbf{O}, \textbf{B-K}, \textbf{I-K} and so on. In order to obtain a categorical distribution, we apply the SoftMax function to $\mathbf{q}^i$ such that
\begin{equation*}
\mathbf{p}^i = \text{SoftMax}(\mathbf{q}^i)
\end{equation*}
where $\mathbf{p}^i$ is the vector of probability estimates serving as a categorical distribution over $\ell$ tags for the word at position $i$. We optimize by computing the standard categorical cross-entropy loss for each of the $n$ individual tag predictions. The final loss to be optimized is the mean over all $n$ individually-computed losses.

A \emph{stacked} Bi-LSTM architecture improves over a single Bi-LSTM architecture given its capacity to learn \emph{deep} contextualized embeddings. While we showed that the stacked approach is better for this particular task in Section~\ref{sec-validation}, it is not necessarily the case that a stacked approach is better in general. We offer an alternative explanation and motivation for using a stacked architecture for this particular problem based on our initial intuition as follows. First, we note that a standalone Bi-LSTM is not able to handle the inference aspect of NER, which entails learning IOB constraints. As an example, in the IOB encoding scheme, it is not possible for a {\bf I-D} tag to immediately follow a {\bf B-E} tag; in this way, the prediction of a tag is directly dependent on the prediction of neighboring tags. This inference aspect is typically handled by a linear-chain CRF. We believe that a stacked Bi-LSTM at least partially handles this aspect in the sense that the first Bi-LSTM (the context encoder) is given the opportunity to form independent preliminary decisions while the second Bi-LSTM is tasked with to making final decisions (based on preliminary ones) that are \emph{more} globally consistent with respect to IOB constraints.

\paragraph{Outcome Objective.}

To predict outcome, we construct a secondary branch in the network path that involves convolving over the word and context embeddings made available in earlier layers. We first define a relation representation $\mathbf{v} \in \mathbb{R}^{d_\text{rel}}$ that is produced by convolving with window sizes 3, 4, and 5 over the context vectors concatenated to entity-bound\footnote{We refer to the process of generating examples for relation classification, wherein mentions of candidate entities in the context are replaced with generic tokens that are also learned during back-propagation, as ``entity-binding.''} versions of the original input; concretely,
\begin{equation*}
\mathbf{v} = \text{CNN}^{[3,4,5]}
\left(  {\begin{array}{*{20}c}
   C_1 \,\, \| \,\, S^{'}_1 \\
   \vdots\\
   C_n \,\, \| \,\, S^{'}_n \\
 \end{array} }  \right).
\end{equation*}
where $S^{'}$ is the entity-bound version of $S$. Based on this outcome representation, we compose two separate softmax outputs: one for PK interactions and one for PD interactions. Concretely, the output layers are 
\begin{equation*}
\mathbf{p}^\text{PK} = \text{SoftMax}(W^{\text{PK}} \mathbf{v} + \mathbf{b}^{\text{PK}})
\end{equation*}
and
\begin{equation*}
\mathbf{p}^\text{PD} = \text{SoftMax}(W^{\text{PD}} \mathbf{v} + \mathbf{b}^{\text{PD}})
\end{equation*}
where $\mathbf{p}^\text{PK} \in \mathbb{R}^{\ell^{\text{PK}}}$ and $\mathbf{p}^\text{PD} \in \mathbb{R}^{\ell^{\text{PD}}}$ are probability estimates serving as a categorical distribution over the outcome label space for PD and PK respectively and $W^{\text{PD}}$, $W^{\text{PK}}$, $\mathbf{b}^{\text{PK}}$, and $\mathbf{b}^{\text{PD}}$ are parameters of the network. For PK, $\ell^{\text{PK}} = 20$ given there are 20 possible NCI Thesaurus codes corresponding to PK outcomes. For PD, $\ell^{\text{PD}} = 2$ as it is a binary classification problem to assess whether the precipitant and effect pair encoded by $S^{'}$ are linked. We optimize using the standard categorical cross-entropy loss on both objectives.

\paragraph{Training Data.} 

In NLM-180, there is no distinction between triggers and effects; moreover, PK effects are limited to coarse-grained (binary) labels corresponding to \emph{increase} or \emph{decrease} in function measurements. Hence, a direct mapping from NLM-180 to Training-22 is impossible. As a compromise, NLM-180 ``triggers'' were mapped to Training-22 \emph{triggers} in the case of unspecified and PK interactions. For PD interactions, we instead mapped NLM-180 ``triggers'' to Training-22 \emph{effects}, which we believe to be appropriate based on our manual analysis of the data. Since we do not have both \emph{trigger} and \emph{effect} for every PD interaction, we opted to ignore trigger mentions altogether in the case of PD interactions to avoid introducing mixed signals.  While trigger recognition has no bearing on relation extraction performance, this policy has the effect of reducing the recall upperbound on NER by about 25\% (more later on upperbound). To overcome the lack of fine-grained annotations for PK outcome in NLM-180, we deploy the well-known bootstrapping approach~\cite{jones1999bootstrapping} to incrementally annotate NLM-180 PK outcomes using Training-22 annotations as seed examples. To mitigate the problem of semantic drift, in each bootstrap cycle, we re-annotated by hand predictions that were not consistent with the original NLM-180 coarse annotations (i.e., active learning~\cite{settles2012active}).

\paragraph{Training Procedure.} 

We train the three objective losses (NER, PK outcome, and PD outcome) in an interleaved fashion at the minibatch~\cite{le2011optimization} level. We use word embeddings of size 200 pre-trained on the PubMed corpus~\cite{pyysalo2013distributional} as input to the network; these are further modified during back-propagation. For the character-level CNN, we set the character embedding size to 24 with 50 filters over a window size of 3; the final character-CNN composition is therefore of length 50. For each Bi-LSTM, the hidden size is set to 100 such that context vectors are 200 in length. For outcome prediction, we used window sizes of 3, 4, and 5 with 50 filters per window size; the final vector representation for outcome prediction is therefore 150 in length.

A held-out development set of 4 drug labels is used for tuning and validation. The models are trained for 30 epochs with check-pointing; only the check-point with the best performance on the development set is kept for testing. We dynamically set the mini-batch size $N_b$ as a function of the number of examples $N$ such that the number of training iterations is roughly 300 per epoch (and also constant regardless of training data size); concretely, $N_b = \lfloor N/300 \rfloor + 1$. As a form of regularization, we apply dropout~\cite{srivastava2014dropout} at a rate of 50\% on the hidden representations immediately after a Bi-LSTM or CNN composition. The outcome objectives are trained such that the gradients of the context encoder weights are downscaled by an order of magnitude (i.e., one tenth) to encourage learning at the later layers. When learning on the NER objective -- the main branch of the network -- the gradients are not downscaled in the same manner. Moreover, when training on the NER objective, we upweight the loss penalty on ``relation'' tags (non-{\bf O} tags) by a factor of 10, which forces the model to prioritize differentiation between different types of interactions over span segmentation. We additionally upweight the loss penalty by a factor of 3 on Training-22 examples compared to NLM-180 examples. We optimize using the Adam~\cite{kingma2015adam} optimization method. These hyper-parameters were tuned during initial experiments.

\section{Results and Discussion}

In this section, we present and discuss the results of our cross-validation experiments. We then describe the ``runs'' that were submitted as challenge entries and present our official challenge results. We discuss these results in Section~\ref{sec-discussion}. 

\subsection{Validation Results}\label{sec-validation}

\begin{table*}[ht]\renewcommand{\arraystretch}{1.3}
      \centering
			\resizebox{\textwidth}{!}{
      \begin{tabular}{@{\extracolsep{1em}}l ccc ccc}
        \toprule
        \, & \multicolumn{3}{c}{Entity (Primary)} & \multicolumn{3}{c}{Relation (Primary)}\\
        \cline{2-4}\cline{5-7}
        \textbf{Model / Data} & \textbf{P} (\%) & \textbf{R} (\%) & \textbf{F} (\%) & \textbf{P} (\%) & \textbf{R} (\%) & \textbf{F} (\%) \\
        \midrule
Stacked Bi-LSTM / Training-22 & 37.22 & 40.74 & 38.90 & 18.76 & 23.50 & 20.86\\
Stacked Bi-LSTM / Training-22 + NLM-180 & 49.45 & 49.79 & 49.62 & 42.54 & 43.56 & 43.05 \\
Char-CNN + Stacked Bi-LSTM / Training-22 + NLM-180 & 51.63 & 50.97 & 51.30 & 43.09 & 44.31 & 43.69\\
Char-CNN + Stacked Bi-LSTM + Tweaks / Training-22 + NLM-180  & 53.72 & 53.76 & \textbf{53.74} & 46.35 & 45.66 & \textbf{46.00}\\
        \midrule
        Char-CNN + Bi-LSTM + TCN / Training-22 + NLM-180 & 48.82 & 50.72 & 49.75 & 39.68 & 41.17 & 39.68\\
        Char-CNN + Stacked TCN / Training-22 + NLM-180 & 41.46 & 48.44 & 44.68 & 32.15 & 36.68 & 34.27\\
        \midrule
        Upperbound due to simplifying assumptions & 99.21 &	74.56	& 85.14 & 97.49 & 81.44 & 88.74\\
        \bottomrule 
      \end{tabular}}
      \caption{Preliminary results based on 11-fold cross validation over Training-22 with two held-out drug labels per fold. When NLM-180 is incorporated, the training data used for each fold consists of 20 non-held out drug labels from Training-22 and all 180 drug labels from NLM-180.}
      \label{tb_validation_results}
    \end{table*}

We present the results of our initial experiments in Table~\ref{tb_validation_results}. Evaluations were produced as as result of 11-fold cross-validation over Training-22 with two drug labels per fold. Instead of macro-averaging over folds, and thereby weighting each fold equally, we evaluate on the union of all 11 test-fold predictions. 

The upperbound in Table~\ref{tb_validation_results} is produced by reducing Training-22 (with gold labels) to our sequence-tagging format and then reverting it back to the original official XML format. Lowered recall is mostly due to simplifying assumptions; e.g., we only consider non-overlapping mentions. For coordinated disjoint cases such as ``X and Y inducers'', we only considered ``Y inducers'' in our simplifying assumption. Imperfect precision is due to discrepancies between the tokenization scheme used by our method and that used to produce gold annotations; this leads to the occasional mismatch in entity offsets during evaluation.

Using a stacked Bi-LSTM trained on the original 22 training examples (Table~\ref{tb_validation_results}; row 1) as our baseline, we make the following observations. Incorporating NLM-180 resulted in a significant boost of more than 20 F1-points in relation extraction performance and more than 10 F1-points in NER performance (Table~\ref{tb_validation_results}; row 2), despite the lowered upperbound on NER recall as mentioned in Section~\ref{sec-method}. Adding character-CNN based word representations improved performance marginally, more so for NER than relation extraction (Table~\ref{tb_validation_results}; row 3). We also implemented several tweaks to the pre-processing and post-processing aspects of the model based on preliminary error analysis including (1) using drug class mentions (e.g., ``diuretics'') as proxies if the drug label is not mentioned directly; (2) removing modifiers such as \emph{moderate}, \emph{strong}, and \emph{potent} so that output conforms to official annotation guidelines; and (3) purging predicted mentions with only stopwords or generic terms such as ``drugs'' or ``agents.'' These tweaks improved performance by more than two F1-points across both metrics (Table~\ref{tb_validation_results}; row 4). 
		
\paragraph{Stacked architecture.} Based on early experiments with simpler models tuned on \emph{relaxed} matching (not shown in Table~\ref{tb_validation_results} and not directly comparable to results displayed in Table~\ref{tb_validation_results}), we found that a stacked Bi-LSTM architecture improves over a single Bi-LSTM by approximately four F1-points on relation extraction (55.59\% vs. 51.55\% F1 tuned on the \emph{relaxed} matching criteria). We moreover found that omitting word embeddings as input at the second Bi-LSTM results in worse performance at 52.91\% F1. 

\paragraph{Temporal Convolution Networks.} We also experimented with using Temporal Convolution Networks (TCNs)~\cite{bai2018empirical} as a ``drop-in'' replacement for Bi-LSTMs. Our attempts involved replacing only the second Bi-LSTM with a TCN (Table~\ref{tb_validation_results}; row 4) as well as replacing both Bi-LSTMs with TCNs (Table~\ref{tb_validation_results}; row 5). The results of these early experiments were not promising and further fine-tuning may be necessary for better performance.

\subsection{Official Test Results}

\begin{table*}[ht]\renewcommand{\arraystretch}{1.3}
  \begin{subtable}{\textwidth}
      \centering
			\resizebox{\textwidth}{!}{
      \begin{tabular}{@{\extracolsep{1em}}ll ccc ccc ccc}
        \toprule
        \, & \, & \multicolumn{3}{c}{Entity (Primary)} & \multicolumn{3}{c}{Relation (Relaxed)} & \multicolumn{3}{c}{Relation (Primary)}\\
        \cline{3-5}\cline{6-8}\cline{9-11}
        \textbf{Training Data} & \textbf{Method} & \textbf{P} (\%) & \textbf{R} (\%) & \textbf{F} (\%) & \textbf{P} (\%) & \textbf{R} (\%) & \textbf{F} (\%) & \textbf{P} (\%) & \textbf{R} (\%) & \textbf{F} (\%) \\
\midrule
Training-22 & Ours (Single model)$^{1}$ & 21.74 & 33.84 & 26.47 & 32.95 & 34.57 & 33.74 & 15.49 & 15.42 & 15.45\\
        \midrule
        \multirow{6}{*}{Training-22 + NLM-180} 
        & Zhang and Kordjamshidi~\cite{zhang2018pe} & 17.00 & 15.86 & 16.41 & - & - & - & - & - & -\\
        & Akhtyamova and Cardiff~\cite{akhtyamova2018extracting} & 37.96 & 20.39 & 26.53 & - & - & - & - & - & - \vspace{0.5em} \\ 
        & Ours (Modifier Coordination) & 28.63 & 27.48 & 28.04 & 38.97 & 30.62 & 34.30 & 21.94 & 16.57 & 18.88\\
        & Ours (Single model) & 29.64 & 31.58 & 30.58 & 38.16 & 33.80 & 35.85 & 21.28 & 18.09 & 19.55\\
        & Dandala et al.~\cite{dandala2018ibm} & 41.94 & 23.19 & 29.87 & 46.60 & 29.78 & 36.34 & 25.24 & 16.10 & 19.66\\
 & Ours (Ensemble) & 29.50 & 37.45 & \textbf{33.00} & 40.55 & 38.36 & \textbf{39.43} & 22.08 & 21.13 & \textbf{21.59}\\
        \midrule
        Training-22 + NLM-180 + HS$^{2}$ & Tang et al.~\cite{tang2018two} & 55.23 & 38.32 & \textbf{45.25} & 71.70 & 45.46 & \textbf{55.64} & 54.43 & 32.76 & \textbf{40.90}\\
        \bottomrule
      \end{tabular}}
      \caption{\vspace{1em}System performance on Official Test Set 1}
      \label{tb_results_test1}
    \end{subtable}
    \begin{subtable}{\textwidth}
      \resizebox{\textwidth}{!}{
      \begin{tabular}{@{\extracolsep{1em}}ll ccc ccc ccc}
        \toprule
        \, & \, & \multicolumn{3}{c}{Entity (Primary)} & \multicolumn{3}{c}{Relation (Relaxed)} & \multicolumn{3}{c}{Relation (Primary)}\\
        \cline{3-5}\cline{6-8}\cline{9-11}
        \textbf{Training Data} & \textbf{Method} & \textbf{P} (\%) & \textbf{R} (\%) & \textbf{F} (\%) & \textbf{P} (\%) & \textbf{R} (\%) & \textbf{F} (\%) & \textbf{P} (\%) & \textbf{R} (\%) & \textbf{F} (\%) \\
\midrule
Training-22 & Ours (Single model)$^{1}$ & 27.77 & 33.31 & 30.29 & 38.38 & 40.85 & 39.58 & 16.17 & 15.39 & 15.77\\
        \midrule
        \multirow{6}{*}{Training-22 + NLM-180} 
        & Zhang and Kordjamshidi~\cite{zhang2018pe} & 17.13 & 21.89 & 19.22 & - & - & - & - & - & -\\
        & Akhtyamova and Cardiff~\cite{akhtyamova2018extracting} & 37.76 & 24.07 & 29.40 & - & - & - & - & - & - \vspace{0.5em} \\ 
        & Ours (Modifier Coordination) & 34.92 & 30.33 & 32.46 & 46.72 & 36.13 & 40.75 & 21.39 & 15.67 & 18.09\\
        & Dandala et al.~\cite{dandala2018ibm} & 44.61 & 29.31 & 35.38 & 50.07 & 36.86 & 42.46 & 22.99 & 16.83 & 19.43\\
        & Ours (Single model) & 35.29 & 33.47 & 34.36 & 44.93 & 37.64 & 40.96 & 22.51 & 17.71 & 19.82\\
 & Ours (Ensemble) & 36.68 & 40.02 & \textbf{38.28} & 49.51 & 44.27 & 
 \textbf{46.74} & 22.53 & 21.13 & \textbf{23.55}\\
        \midrule
        Training-22 + NLM-180 + HS$^{2}$ & Tang et al.~\cite{tang2018two} & 51.23 & 42.39 & \textbf{46.39} & 66.99 & 49.58 & \textbf{56.98} & 48.92 & 34.49 & \textbf{40.46}\\
        \bottomrule
      \end{tabular}}
      \caption{System performance on Official Test Set 2}
      \label{tb_results_test2}
    \end{subtable}
      \caption{Comparison of our method with that of other teams in the top 5. Only the best performing method of each team is shown; methods are grouped by available training data and ranked in ascending order by relation extraction (primary) performance followed by entity recognition performance.}
      \footnotesize{$^{1}$This model was not submitted and is shown for reference only} \\
      \footnotesize{$^{2}$HS refers to a private dataset of 1148 sentences manually-annotated by Tang~et~al.~\cite{tang2018two} according to official guidelines}
      \label{tb_results}
    \end{table*}
 
Our final system submission is based on a stacked Bi-LSTM network with character-CNNs trained on both Training-22 and NLM-180 (corresponding to row 4 of Table~\ref{tb_validation_results}). We submitted the following three runs based on this architecture:
\begin{enumerate}
\item A single model.\vspace{-0.1em}
\item An ensemble over \emph{ten} models each trained with randomly initialized weights and a random development split. Intuitively, models collectively ``vote'' on predicted annotations that are kept and annotations that are discarded. A unique annotation (entity or relation) has one vote for each time it appears in one of the \emph{ten} model prediction sets. In terms of implementation, unique annotations are incrementally added (to the final prediction set) in order of descending vote count; subsequent annotations that conflict (i.e., overlap based on character offsets) with existing annotations are discarded. Hence, we loosely refer to this approach as ``voting-based'' ensembling.\vspace{-0.1em}
\item A single model with pre/post-processing rules to handle modifier coordinations; for example, ``X and Y inducers'' would be correctly identified as two distinct entities corresponding to ``X inducers'' and ``Y inducers.'' Here, we essentially encoded ``X and Y inducers'' as a single entity when training the NER objective; during test time, we use simple rules based on pattern matching to split the joint ``entity'' into its constituents.\vspace{-0.1em}
\end{enumerate}

Eight teams participated in task 1 while four teams participated in task 2. We record the relative performance of our system (among others in the top 5) on the two official test sets in Table~\ref{tb_results}. For each team, we only display the performance of the best run for a particular test set. Methods are grouped by the data used for training and ranked in ascending order of \emph{primary} relation extraction performance followed by entity recognition performance. We also included a single model trained solely on Training-22, that was not submitted, for comparison. Our voting-based ensemble performed best among the three systems submitted by our team on both NER and relation extraction. In the official challenge, this model placed \emph{second} overall on both NER and relation extraction. 

Tang et al.~\cite{tang2018two} boasts the top performing system on both tasks. In addition to Training-22 and NLM-180, the team trained and validated their models on a set of 1148 sentences sampled from DailyMed labels that were manually annotated according to official annotation guidelines. Hence, strictly speaking, their method is not directly comparable to ours given the significant difference in available training data.

\subsection{Discussion}\label{sec-discussion}

 While precision was similar between the three systems (with exceptions), we observed that our ensemble-based system benefited mostly from improved recall. This aligns with our initial expectation (based on prior experience with deep learning models) that an ensemble-based approach would improve stability and accuracy with deep neural models. Although including NLM-180 as training data resulted in significant performance gains during 11-fold cross validation, we find that the same  improvements were not as dramatic on either test sets despite the 800\% gain in training data. As such, we offer the following analysis. First, we suspect that there may be a semantic or annotation drift between these datasets as annotation guidelines evolve over time and as annotators become more experienced. To our knowledge, the datasets were annotated in the following order: NLM-180, Training-22, and finally Test Sets 1 and 2; moreover, Test Sets 1 and 2 were annotated by separate groups of annotators. Second, having few but higher quality examples may be more advantageous than having many but lower quality examples, at least for this particular task where evaluation is based on matching exact character offsets. Finally, we note that the top performing system exhibits superior performance on Test Set 1 compared to Test Set 2; interestingly, we observe an inverse of the scenario in our own system. This may be an indicator that our system struggles with data that is more ``sparse'' (as previously defined in Section~\ref{sec-dataset}).

\section{Conclusion}

We presented a method for jointly extracting precipitants and their interaction types as part of a multi-task framework that additionally detects interaction outcome. Among three ``runs'', a ten model voting-ensemble was our best performer. In future efforts, we will experiment with Graph Convolution Networks~\cite{zhang2018graph} over dependency trees as a ``drop-in'' replace for Bi-LSTMs to assess its suitability for this task.

\section*{Acknowledgements}

This research was conducted during TT's participation in the Lister Hill National Center for Biomedical Communications (LHNCBC) Research Program in Medical Informatics for Graduate students at the U.S. National Library of Medicine, National Institutes of Health. HK is supported by the intramural research program at the U.S. National Library of Medicine, National Institutes of Health. RK and TT are also supported by the U.S. National Library of Medicine through grant R21LM012274.

\bibliographystyle{unsrt}
\bibliography{scoonerdb}

\end{document}